%% file: main.tex
\documentclass[letterpaper, 10pt, conference]{ieeeconf}  

\IEEEoverridecommandlockouts                              

\usepackage{graphics} 
\usepackage{epsfig} 
\usepackage{times} 
\usepackage{amsmath} 
\usepackage{amssymb}  
\usepackage[bookmarks=true]{hyperref}
\usepackage{xcolor,colortbl}
\usepackage{color}
\usepackage[ruled,vlined]{algorithm2e}
\usepackage{siunitx}
\usepackage{multirow}
\usepackage[skip=0.3\baselineskip]{caption}
\usepackage{booktabs}
\usepackage{cite}
\usepackage{balance}

\newcommand{\bs}{\boldsymbol}

\definecolor{Gray}{gray}{0.85}
\definecolor{LightCyan}{rgb}{0.88,1,1}
\definecolor{somegray}{rgb}{0.5, 0.5, 0.5}
\newcommand{\darkgrayed}[1]{\textcolor{somegray}{#1}}
\makeatletter

\newcommand*\titleheader[1]{\gdef\@titleheader{#1}}
\AtBeginDocument{%
  \let\st@red@title\@title
  \def\@title{%
    \vskip-4.2em
    \bgroup\normalfont\large\centering\@titleheader\par\egroup
    \vskip0.9em\st@red@title}
}
\makeatother

\titleheader{\darkgrayed{This paper has been accepted for publication at the
\\ IEEE International Conference on Robotics and Automation (ICRA), Xi’an, 2021. 
\copyright IEEE} }

\title{\LARGE \bf
Autonomous Overtaking in Gran Turismo Sport \\
Using Curriculum Reinforcement Learning 
}

\author{
    Yunlong Song$^{\ast}$,
    HaoChih Lin$^{\ast}$,
    Elia Kaufmann,
    Peter D\"urr,
    and Davide Scaramuzza 
    \thanks{    
    $^{\ast}$These two authors contributed equally.
    Y. Song, E. Kaufmann, and D. Scaramuzza are with the Robotics and Perception Group, Dep. of Informatics, University of Zurich, and Dep. of Neuroinformatics, University of Zurich and ETH Zurich, Switzerland (\protect\url{http://rpg.ifi.uzh.ch}). H. Lin and P. D\"urr are with Sony AI Zurich. This work was supported by Sony AI Zurich, the National Centre of Competence in Research (NCCR) Robotics through the Swiss National Science Foundation, and the European Research Council Consolidator Grant (ERC-CoG) under the European Union’s Horizon 2020 Research and Innovation Programme (Grant agreement No. 864042).
    }
}

\begin{document}

\maketitle
\thispagestyle{empty}
\pagestyle{empty}

\input{sections/abstract}

~\\
\textbf{Video:} \url{https://youtu.be/e8TVPv4D4O0}

\input{sections/introduction}
\input{sections/relatedwork}
\input{sections/methodology}

\input{sections/experiments}

\input{sections/conclusion}
\input{sections/acknowledgments}
\newpage

\balance
\bibliographystyle{IEEEtran}
\bibliography{ref}

\end{document}

%% file: sections/abstract.tex
\begin{abstract}
Professional race-car drivers can execute extreme overtaking maneuvers. 
However, existing algorithms for autonomous overtaking either rely on simplified assumptions about the vehicle dynamics or try to solve expensive trajectory-optimization problems online.
When the vehicle approaches its physical limits, existing model-based controllers struggle to handle highly nonlinear dynamics, and cannot leverage the large volume of data generated by simulation or real-world driving. 
To circumvent these limitations, we propose a new learning-based method to tackle the autonomous overtaking problem.
We evaluate our approach in the popular car racing game  Gran Turismo Sport, which is known for its detailed modeling of various cars and tracks.
By leveraging curriculum learning, our approach leads to faster convergence as well as increased performance compared to vanilla reinforcement learning.  
As a result, the trained controller outperforms the built-in model-based game AI and achieves comparable overtaking performance with an experienced human driver. 
\end{abstract}

%% file: sections/introduction.tex
\section{INTRODUCTION}
The goal of autonomous overtaking in car racing is to overtake 
the opponents as fast as possible while avoiding collisions.
Experienced race-car drivers can operate a vehicle at the limits of handling and, 
at the same time, perform overtaking during very extreme maneuvers. 
Developing an autonomous system that can achieve the same level of human control performance, or even go beyond, could not only shorten the travel time and
reduce transportation costs but also avoid fatal accidents. 

However, developing such an autonomous overtaking system is very challenging for several reasons:
1) The entire system, including the vehicle, the tire model, and the vehicle-road interaction,
has highly complex nonlinear dynamics. 
2) The intentions of other opponents are unknown, rendering most high-level trajectory planning algorithms 
incapable of reliably generating accurate overtaking trajectories. 
3) The vehicle is already close to its physical limits, leaving very limited control authority for
executing overtaking maneuvers. 
\begin{figure}[!htp]
    \centering
    \includegraphics[width=0.48\textwidth]{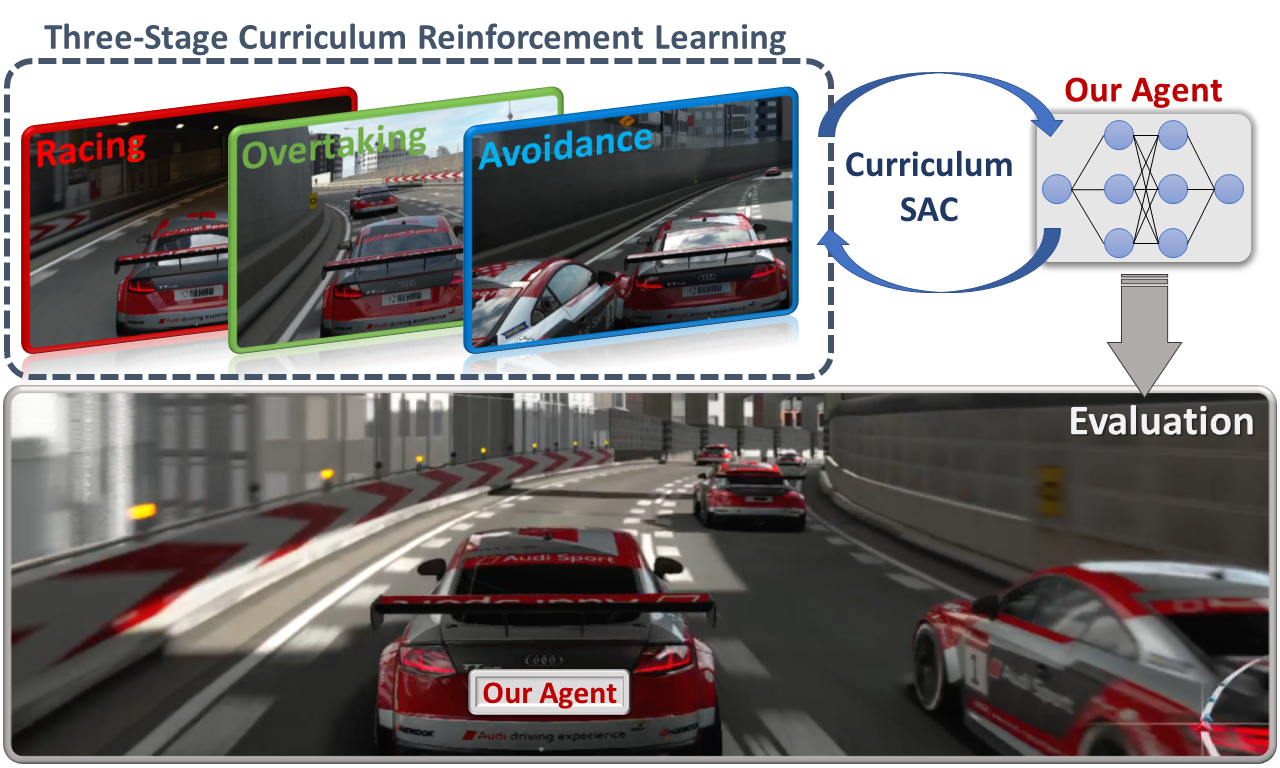}
    \caption{A system overview of the proposed curriculum reinforcement learning method
    for addressing the autonomous overtaking problem in Gran Turismo Sport.}
     \label{fig:racing_env}
\end{figure}

Previous methods tackled the problem using classical 
trajectory generation and tracking techniques and relied on tools from dynamic modeling, optimal control, and nonlinear programming. 
Despite all the successes~\cite{dixit2018trajectory}, this line of research has several limitations.
For example, many trajectory planning algorithms~\cite{paden2016survey, buyval2017deriving, heilmeier2019minimum} use a 
simplified vehicle model and neglect several real-world effects, such as the tire-road interaction and aerodynamic effects.
These algorithms escalate in complexity when considering high-fidelity vehicle models and complex interactions
among vehicles.  

Recently, deep reinforcement learning~(RL) has emerged as an effective approach in solving
complex robotic control problems~\cite{lee2020learning, fuchs2020super, jaritz2018end}. 
Particularly, model-free deep RL trains a parametric
policy by directly interacting with the environment and does
not assume knowledge of an exact mathematical model of the
system, making the method well-suited for highly nonlinear
systems and complex tasks.
Furthermore, the neural network policies allow flexible controller design, allowing
different state representations that range from high-dimensional images to low-dimensional states. 

Here, we present a new learning-based system for high-speed autonomous overtaking. 
The key is to leverage task-specific curriculum RL
and a novel reward formulation to train an end-to-end neural network controller. 
Our approach manifests faster convergence as well as increased performance
compared to vanilla deep RL, which instead trains neural network policies
directly for overtaking without any prior knowledge about driving. 
The proposed curriculum learning procedure can transfer the knowledge that is obtained from a single-car racing task to solve more complicated overtaking problems. 
As a result, our trained controller outperforms the built-in model-based controller and achieves comparable overtaking performance with an experienced human driver.

%% file: sections/relatedwork.tex
\section{Related Work}
A bulk of research in autonomous driving has been
focusing on developing safe overtaking systems in low-speed driving scenarios~\cite{dixit2018trajectory}. 
We categorize prior work in the domain of autonomous overtaking into two groups: 
model-based approaches and learning-based approaches.

\subsubsection{Model-based}
Model-based approaches attempt to tackle the problem via a modular architecture design, which
breaks the overtaking problem down into trajectory planning and trajectory tracking. 
For instance, sampling-based trajectory planning methods~\cite{ma2014fast, kuwata2008motion}, such as
Rapidly Exploring Random Trees~(RRT), have been proposed for planning safe trajectories
for autonomous overtaking. 
These methods generally make use of simplified vehicle models and basic vehicle kinematics. 
However, when a car operates close to the limits of handling and drives at high-speed, 
it is insufficient to ignore many real-world effects, such as tire-road interaction 
and aerodynamics effects introduced by the motion of other vehicles.  

Optimization-based approaches, such as Model Predictive Control~(MPC), are effective
solutions to trajectory planning and tracking in autonomous overtaking~\cite{dixit2019trajectory, wang2019game, petrov2014modeling}, 
thanks to their capability of handling different constraints and robust performance against disturbances.
Similar to motion planning algorithms, optimization-based approaches rely on the assumption of a simplified car model, and thus,
do not guarantee that they can handle very complex nonlinear system dynamics. 
Besides, the requirement of solving nonlinear optimization online is computationally demanding
for embedded systems. 

\subsubsection{Learning-based} 
Learning-based approaches, such as imitation learning~\cite{farag2018behavior, pomerleau1991efficient}
and reinforcement learning~\cite{schwartingdeep, li2015reinforcement, loiacono2010learning},
can in principle address the limitations of traditional modular and model-based approaches 
by learning parameterized policies that directly map sensory observations to control commands. 
One of the key advantages of using learning-based approaches is that they do not require perfect knowledge 
about the vehicle and its environment. 

While imitation learning is an effective approach for training a neural network policy 
using experienced data demonstrated by human experts, overtaking is a sparse signal 
and can be difficult for human drivers. 
When deployed naively, imitation learning is sensitive to the distribution shift between the observations induced by the expert policy and the network policy. 
This problem can be alleviated using DAGGER~\cite{ross2011reduction}, a time-consuming and expensive process for data collection.

Reinforcement learning (RL) seems to offer real potential for solving such complex decision-making problems
by maximizing a reward signal that can formulate the overtaking problem properly. 
However, most advances in RL published to date are largely empirical. 
A thorough study on training methods and design choices in the autonomous overtaking domain
is, unfortunately, lacking in the community. 
Our work is inspired by~\cite{fuchs2020super}, but extends it to the more complex 
and challenging overtaking domain. 

%% file: sections/methodology.tex
\section{METHODOLOGY}
This section introduces the problem formulation of autonomous overtaking and its reward function design and describes how curriculum can be combined with off-policy actor-critic methods for training the neural network policy. 

\begin{figure*}[!htp]
    \includegraphics[width=1.0\textwidth]{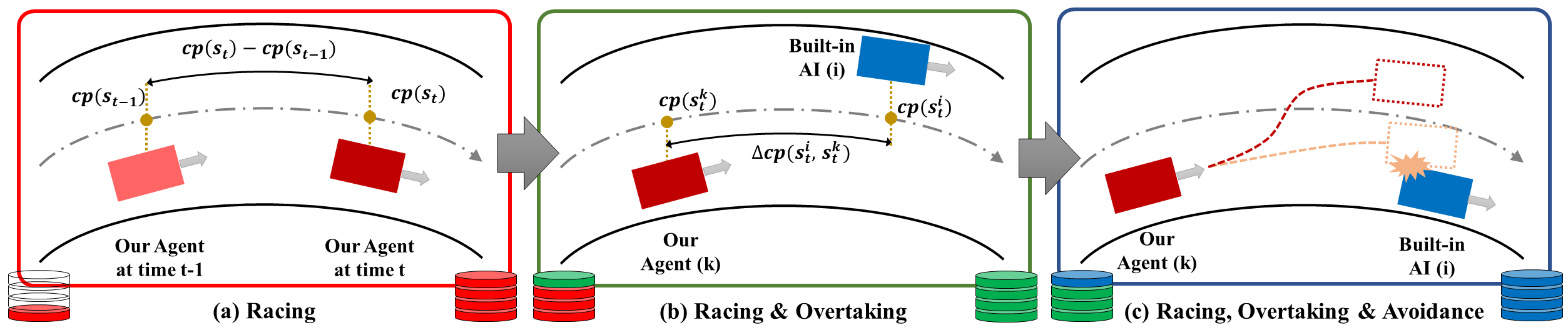}
    \caption{
    An illustration of the proposed three-stage curriculum reinforcement learning for autonomous race car overtaking. 
    }
    \label{fig:curriculum_learning}
\end{figure*}

\subsection{Problem Formulation}
High-speed overtaking in car racing involves two main objectives: minimizing 
the total overtaking time and avoiding collisions between the agent and other vehicles or obstacles.   
Intuitively, the agent that takes a short period to overtake its opponents needs to drive at high speed and 
has high collision probability, and vice versa. 
Hence, the optimal overtaking strategy defines the best trade-off between these two 
competing objectives. 

\subsubsection{The Racing Problem}
We first consider a single-player racing problem, in which the goal is to drive a race car 
on a given track in minimum time.
Instead of minimizing the time directly, the time-optimal objective is normally
reformulated as minimizing the path of least curvature or the shortest path in order 
to use numerical optimization methods~\cite{verschueren2014towards}.
In~\cite{fuchs2020super}, the authors propose a course-progress-proxy reward formulation,
which closely represents the lap time and can be maximized using reinforcement learning. 
The course progress is determined by projecting the car’s position to the point along
the centerline (see Fig.~\ref{fig:curriculum_learning}). 
Hence, the racing reward $r_t^\text{racing}$ at the time stage~$t$ is defined as:
\begin{equation} 
    \label{eq:racing_reward}
    r^\text{racing}_{t} = \left( cp(\bs{s}_t) - cp(\bs{s}_{t-1}) \right) - c_w\rho_w|\bs{v}_t|^2 
\end{equation}
where $cp(\bs{s}_t)$ is the centerline projection based on current car position $\bs{s}_t$.
Here, $\rho_w$ is a binary flag indicating whether or not the wall collision occurs,
and $c_w\geq0$ is a hyperparameter for weighting the wall collision penalty.
From the physical point of view, the first two terms encourage the learning policy 
to drive as fast as possible and the last term incentivizes to avoid the wall collision in the meantime. 
The last term is also depending on the collision kinetic, 
which is proportional to the square of the car's speed ($\bs{v}_t$).

\subsubsection{The Overtaking Problem}
The overtaking problem in car racing includes not only the objective of minimizing lap time,
but also, avoiding collisions with other vehicles. 
We propose a novel continuous reward function ($r_{t}^\text{overtaking}$) for the overtaking problem. 
The formulation of the proposed continuous reward is expressed as: 
\begin{equation}
\label{eq:overtaking_reward}
\begin{aligned}
& r_{t}^\text{overtaking} =  r_{t}^\text{racing}  - c_c\rho_c|\bs{v}_t|^2 +  \\ 
& \sum_{\forall i \in C \setminus \{k\}} \left\{\rho^i  c_r [\Delta cp(s^i_{t-1}, s^k_{t-1})-\Delta cp(s^i_t, s^k_t)]\right\}
\end{aligned}
\end{equation}
where,
\begin{align*}
    \Delta cp(s^i_t, s^k_t) & = cp(s^i_t) - cp(s^k_t) \\
    \rho^i = \rho(s^i_t, s^k_t) & = \begin{cases}
            1, & |\Delta cp(s^i_t, s^k_t)| < c_d \\ 
            0, & \text{Otherwise} 
          \end{cases}
\end{align*}
where $C$ is a set of total simulated cars on the track, $k$ represents the ego-car 
controlled by the learning policy, $c_d$ is a hyperparameter for the detection range, 
$c_r$ is a hyperparameter that trades off between the aggressiveness of the overtaking maneuver and 
the collision penalty.
Here, $\rho_c$ is a binary flag indicating whether or not a car collision occurs, 
and $c_c\geq0$ weights the car collision penalty.

The idea behind using the subtraction~($\Delta cp(s^i_{t-1}, s^k_{t-1})-\Delta cp(s^i_{t}, s^k_{t})$) inside the summation symbol in Eq.~(\ref{eq:overtaking_reward}) is 
to continuously encourage our agent ($k$) to approach the front opponent vehicle~($i$) when it is driving behind,
while keep maximizing the relative distance once it has overtaken the opponent vehicle~($i$). 
The illustration of the proposed idea is depicted in Fig.~\ref{fig:curriculum_learning}. 
By maximizing the proposed overtaking reward, our agent can learn to 
perform overtaking maneuvers and collision avoidance.  

\subsection{Observation and Action}
The definition of both the observation space and the action space is described in Table \ref{tab:observation-action-space}. 
We denote the observation vector as $\bs{o}_t = [\bs{v}_t, \Dot{\bs{v}}_t, \theta, \bs{d}_t, \delta_{t-1}, f_t, f_c, \bs{c}_L]$,
and use them as the input to our neural networks.
The input normalization is important in most learning algorithms since different features might 
have totally different scales.
We apply $z$-score normalization to all features in the observation vector, 
except for the 2D Lidar measurements $\bs{d}_t$, which is normalized using min-max normalization.
We compute the $z$-score normalization using sampled states from the environment. 
The control actions are the steering angle and the combined throttle and brake signal, 
denoted as~$[\delta_t, \omega_t]$ separately.

The 2D Lidar distance measurements detect the relative distance between our agent and other objects, such as 
other vehicles and the wall.
We use a distance vector $\mathbf{d}_t \in \mathbb{R}^{72}_{>0}$ obtained from a set of 72
equally spaced Lidar beams with a maximum detection range of 20~\si{\meter} 
arranged between $-108\si{\degree}\sim108\si{\degree}$ in front of vehicle. 
We constrain both the field-of-view and the detection range for a fair comparison to the human driver.


\input{tables/obs_act}

\subsection{Curriculum Soft Actor-Critic}
We use Soft Actor-Critic~(SAC) for training a neural network policy
that can maximize the overtaking reward. 
However, like most off-policy algorithms, SAC suffers from ``extrapolation error", 
a phenomenon in which unseen state-action pairs are 
erroneously estimated to have unrealistic values~\cite{fujimoto2019off}. 
For example, in the overtaking task, it might take the agent many explorations in order to see
a single overtaking since it does know how to drive at the early training stage.
Hence, maximizing the overtaking reward directly leads to premature convergence and results in
poor final policy. 

\subsubsection{Three-stage Curriculum Reinforcement Learning}
A key ingredient to address this problem for a complex environment is to use curriculum learning. 
In particular, we combine SAC with a 3-stage curriculum learning procedure.
In stage one, we train a policy (with random weights) for high-speed racing.
We use a randomly initialized neural network and train it for the single-player racing (without overtaking)
by maximizing the racing reward function~(Eq.~(\ref{eq:racing_reward})).
We stop the training when the agent is capable of driving the car at a very high speed.
In stage two, we continuously train the same policy for aggressive racing and overtaking.
We load the pre-trained policy done in the first stage and reconfigure the racing environment by adding an extra
vehicle, which is controlled by the built-in game AI controller.
We initialize the distance between our agent and the build-in agent with 200 meters separation along the centerline. 
Before training, it is important to keep the old replay buffer, and reinitialize the weights of  
the exploration term in the stochastic policy in that the policy maintains sufficient explorations. 
We update the policy by maximizing the overtaking reward~(Eq.~\ref{eq:overtaking_reward}) and using new sampled trajectories.
In stage three, we obtain a final policy that can race the car at high speed, overtake its opponents,
and avoid collisions.
This is achieved by increasing the penalty term in the overtaking reward and
training the policy with new samples. 
It is important to use a fixed size first-in-first-out replay buffer, since the racing 
data will be replaced gradually with new overtaking samples. 

\subsubsection{Distributed Sampling Strategy}
The second key ingredient in achieving better global convergence 
for complex environments is to use a distributed sampling strategy for the data collection.
Similar to~\cite{fuchs2020super}, we use a distributed sampling strategy, 
in which we use multiple simulators~(4) in parallel, each 
simulating multiple cars~(20) on the same racing track. 
In other words, we can achieve $4 \times 20$ faster sampling speed than
using a single racing environment. 
Most importantly, the sampled trajectories cover most of the track segments,
and led to a dataset that highly correlates with the true state-action distribution.
As a result, we achieve fast data collection and stable policy training. 
We use this sampling strategy throughout all training stages.

%% file: tables/obs_act.tex
\begin{table}[h]
\caption{The observation space and the action space.}
\label{tab:observation-action-space}
\begin{center}
\begin{tabular}{c||c|c}
\toprule
\rowcolor{Gray}
\multicolumn{3}{c}{Observation Space ($\mathbb{R}^{96}$)}\\
\midrule
$\bs{v}_t$ & linear velocity in body frame & $\mathbb{R}^3$  \\
\midrule
$\Dot{\bs{v}}_t$  & linear acceleration in body frame & $\mathbb{R}^3$ \\
\midrule
$\theta$  & angle between heading and tangent to the centerline & $\mathbb{R}$\\
\midrule
$\bs{d}_t$ & 2D Lidar measurements (-108\si{\degree} $\sim$108\si{\degree}, 20\si{\meter}) & $\mathbb{R}^{72}_{> 0}$ \\
\midrule
$\delta_{t-1}$ & previous steering command & $\mathbb{R}$ \\
\midrule
$f_t$ & binary flag with 1 indicating wall collision & $\mathbb{R}$ \\
\midrule
$f_c$ & binary flag with 1 indicating car collision & $\mathbb{R}$ \\
\midrule
$\bs{c}_L$ & looking forward curvature of centerline (0.2$\sim$3.0 sec) & $\mathbb{R}^{14}$\\
\toprule
\rowcolor{Gray}
\multicolumn{3}{c}{Action Space ($\mathbb{R}^{2}$)}\\
\midrule
$\delta_t$ & steering angle in rad, $\delta_t \in [-\pi/6, \pi/6]$ & $\mathbb{R}$\\
\midrule
$w_t$ & combination of throttle ($w_t>0$) and brake ($w_t<0$) & $\mathbb{R}$\\
\bottomrule
\end{tabular}
\end{center}
\end{table}

%% file: sections/experiments.tex
\section{EXPERIMENTS}
We design experiments to answer the following research questions:
\begin{itemize}
    \item Can our curriculum learning speed up training and improve sample efficiency in comparison with standard training~(Section~\ref{subsec:curriculum_training})?
    \item How should we evaluate the overtaking performance~(Section~\ref{subsec:evaluation})?
    \item What are the overtaking strategies learned by our approach?~(Section~\ref{subsec:learned_overtakin})?
\end{itemize} 

\subsection{Experimental Setup}
\label{subsec:exp_setup}
We conduct our experiment using Gran Turismo Sport~(GTS).  
We train our algorithm on a desktop with an i7-8700 CPU and a GTX 1080Ti GPU.
We use a custom implementation~\cite{fuchs2020super} of the Soft Actor-Critic algorithm 
that is based on the open-source baselines~\cite{SpinningUp2018}. 
GTS runs on a PlayStation~4, and we interact with GTS using an Ethernet connection.
We treat GTS as a black-box simulator since we do not have direct access
to the vehicle dynamics and the environment in GTS.
We choose ``Audi TT Cup 16" as the simulated car model and ``Tokyo Expressway - Central Outer Loop" as the race track. 
The hyperparameters of SAC for the training are listed in TABLE~\ref{tab-training-hyperparameters}.
\input{tables/training_hyperparameters} 

 
\subsection{Curriculum Policy Training for Overtaking}
\label{subsec:curriculum_training}
The first step towards autonomous overtaking in car racing is to obtain a policy that can drive faster than the opponents (the built-in AI)
in a single-car racing environment. 
The built-AI uses a rule-based approach to follow a predefined trajectory, similar to~\cite{ni2019robust, hellstrom2006follow}.
We study different approaches to obtain such a policy before learning to overtake, including naive Behavior Cloning (BC)~\cite{pomerleau1989alvinn}, Generative Adversarial Imitation Learning~(GAIL)~\cite{ho2016generative}, 
Deep Planning Network~(PlaNet)~\cite{hafner2019learning}, 
and Twin Delayed Deep Deterministic Policy Gradient (TD3)~\cite{fujimoto2018addressing}.  
We use the same neural network structure and observation representation for all methods. 
The training data required by imitation learning is collected 
using demonstrations from both the built-AI and human experts. 
The experimental result is shown in TABLE~\ref{tab-baseline-comparison},
only the policy trained using the model-free RL (SAC and TD3) can outperform the built-in AI in
the single-car time trial race. 
\input{tables/baseline_comparison_result}

\begin{figure}[t]
    \centering
    \includegraphics[width=0.5\textwidth]{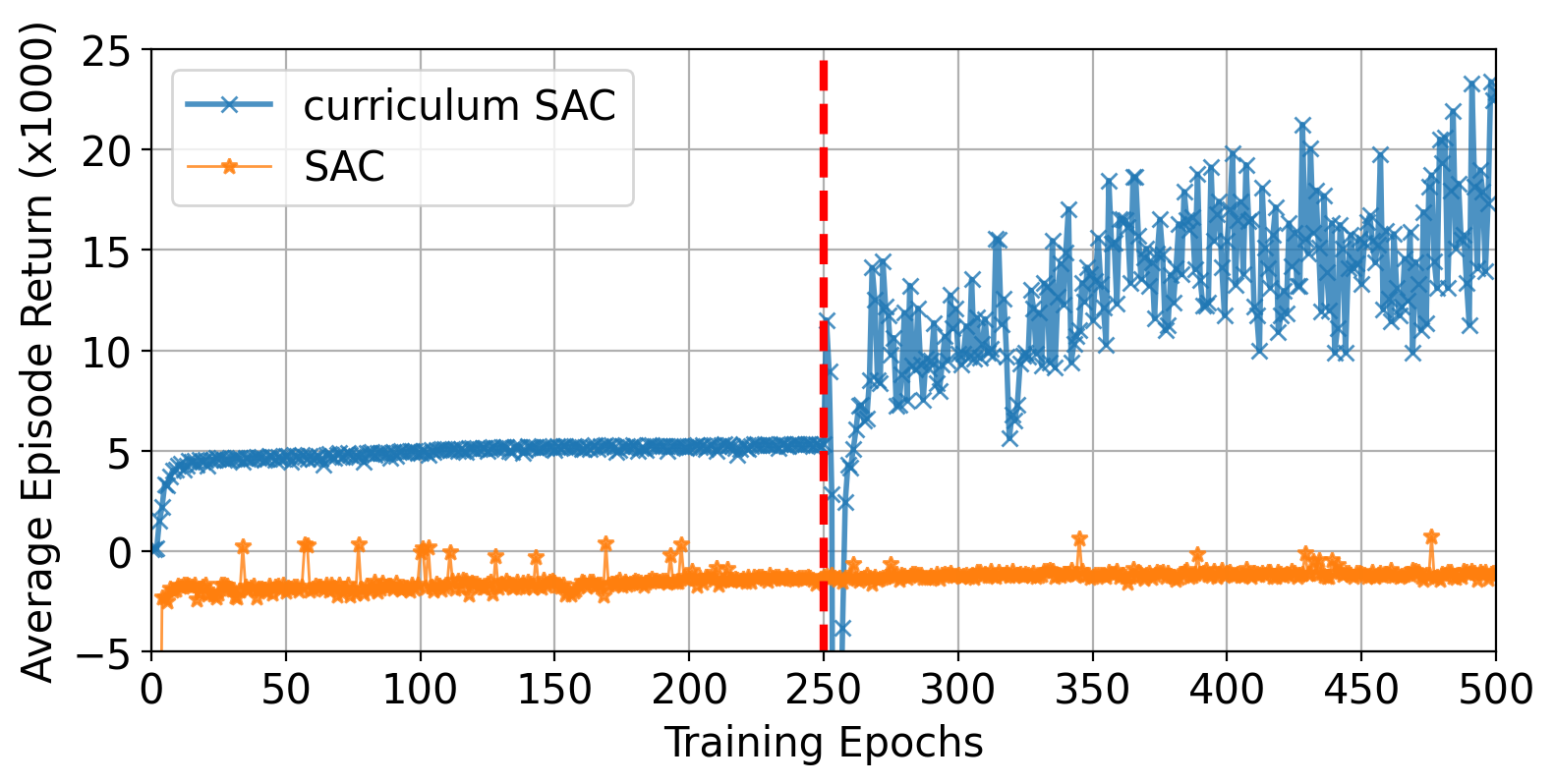}
    \caption{A comparison of the learning curves using 
    different training methods. The red dash line in the middle represents the switch from stage one to stage two. 
    }
     \label{fig:learning_curve}
\end{figure}

To understand the effect of the proposed curriculum learning on policy training, 
we compare the training curves of curriculum SAC with standard SAC.
For standard SAC, we train neural network policies 
by directly maximizing the overtaking reward~(Eq.~(\ref{eq:overtaking_reward})). 
By contrast, for curriculum SAC, we first design a single-player racing environment and 
train a neural network policy by maximizing the racing reward~(Eq.~(\ref{eq:racing_reward})).
Then, we configure an overtaking environment and 
continuously train the policy by maximizing the overtaking reward~(Eq.~(\ref{eq:overtaking_reward}))
with collision weights of $c_w = c_c= 0.005$.
The learning curves are shown in Fig.~\ref{fig:learning_curve}. 
As a result, the proposed curriculum SAC outperforms standard SAC in terms of
sample efficiency and final policy performance. 

\subsection{Evaluation of the Overtaking Performance}
\label{subsec:evaluation}
Evaluating the overtaking performance can be complicated as there are multiple metrics,
such as the total travel time or the total collision time. 
These values are generally difficult to obtain in the real world.  
GTS provides precise quantitative measurements of those metrics.
We propose four objective evaluation metrics for evaluating the obtained policy: 
1) total travel time, 2) total travel distance, 3) total car collision time,
and 4) total wall collision time.

We train three different agents using the overtaking reward with different hyperparameters
and different training procedures for benchmark comparisons.
For Agent1, we use only the first 2 stages that include single-car racing
and multiple-car overtaking.
For Agent2, we use 3-stage training, where the second and the third stage has same collision weights of $c_w = c_c= 0.005$.
For Agent3, we use 3-stage training, where the third stage has larger collision weights of $c_w = c_c= 0.01$ than the second stage, 
which has collision weights of $c_w = c_c= 0.005$. 

To evaluate the overtaking performance of 3 trained agents, 
we use two different settings for the evaluation experiment.
We place 5 opponent vehicles in front of the trained agent with an initial separation distance 
of \SI{50}{\meter} (setting A) and of \SI{200}{\meter} (setting B).
In addition, we invite an expert player TG (name omitted for reasons of anonymity) as a human baseline.
Both the human player and our agent use exactly the same settings and have to overtake all the 5 opponent vehicles. 

\begin{figure}[t]
    \centering
    \includegraphics[width=0.48\textwidth]{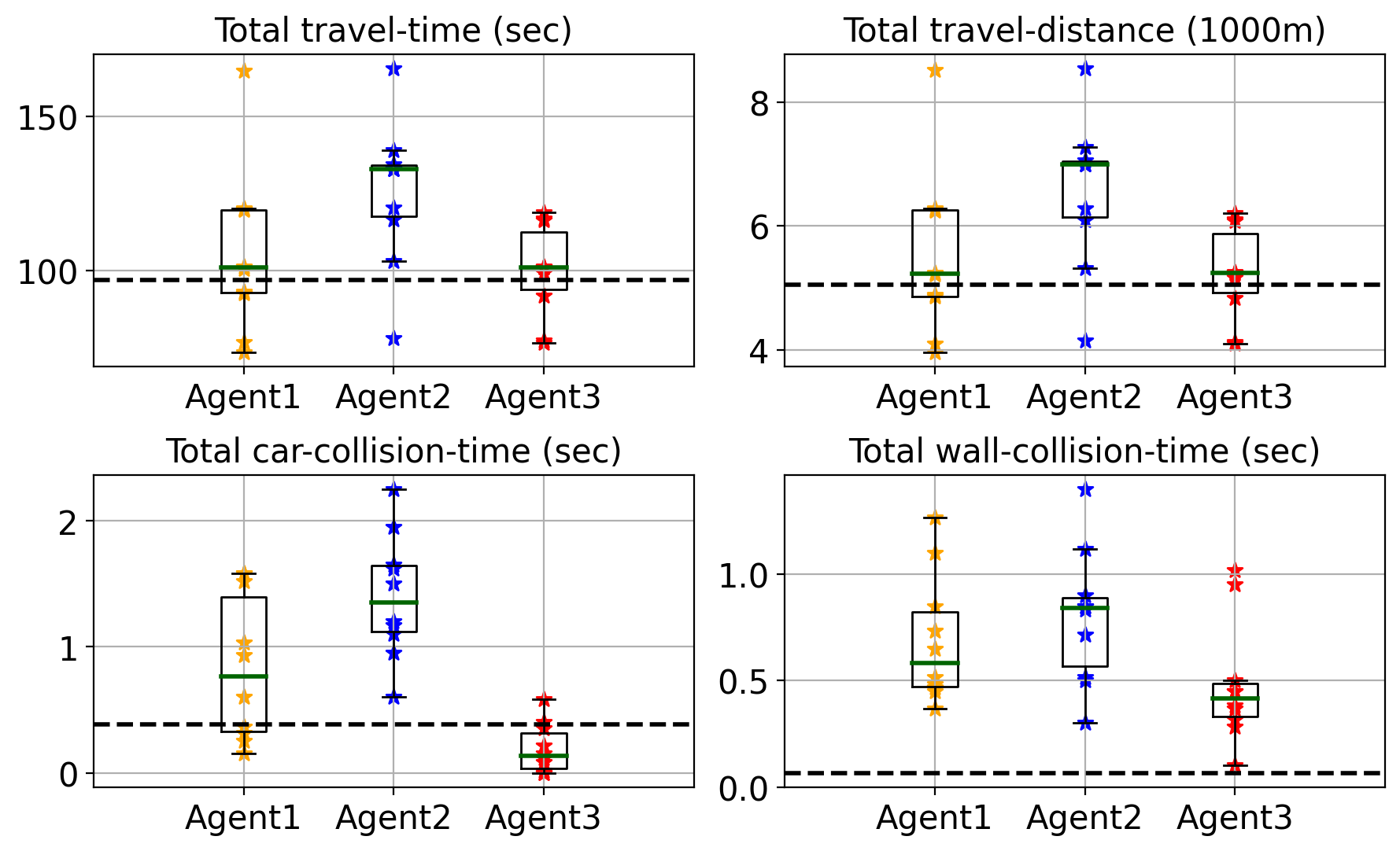}
    \caption{Evaluation comparisons for the setting A. 
    The dashed lines indicate the human player's best performance.}
     \label{fig:setting_A}
\end{figure}

\begin{figure}[t]
    \centering
    \includegraphics[width=0.48\textwidth]{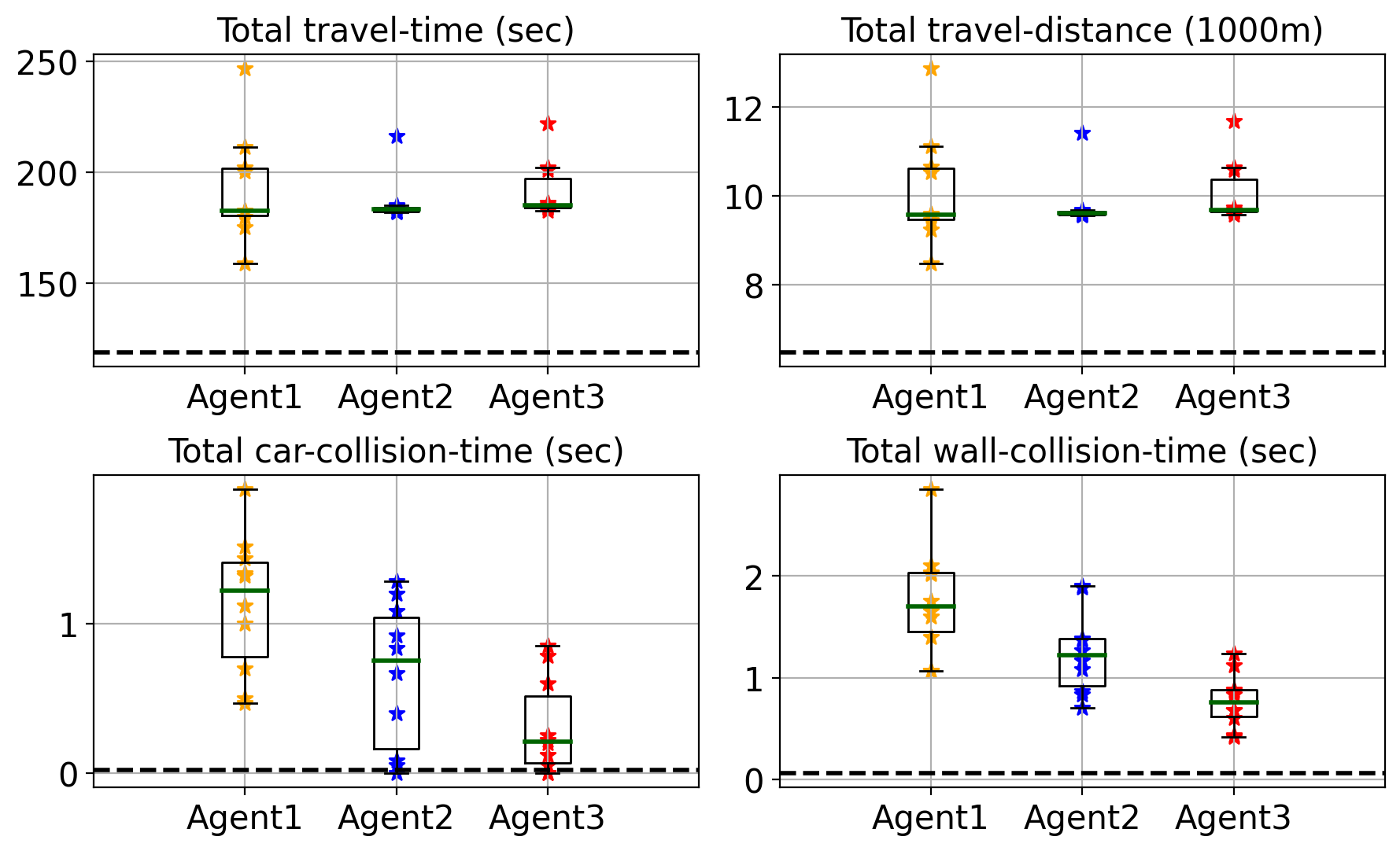}
    \caption{Evaluation comparisons for the setting B.  
    The dashed lines indicate the human player's best performance.}
     \label{fig:setting_B}
\end{figure}

We compute the evaluation metrics for each trained agent 
by conducting the experiment repeatedly 10 times, and for the human player
by repeating the same experiment 2 times. 
We take the best result from the human player as our baseline.
The evaluation results are shown in Fig.~\ref{fig:setting_A} and Fig.~\ref{fig:setting_B}.
Both our agents and the human player are capable of overtaking all 5 opponent vehicles. 
Our agents achieve comparable overtaking performance as the human expert in setting A. 

\subsection{Learned Overtaking Behaviors}
\label{subsec:learned_overtakin}

\begin{figure*}[!htp]
    \includegraphics[width=1.0\textwidth]{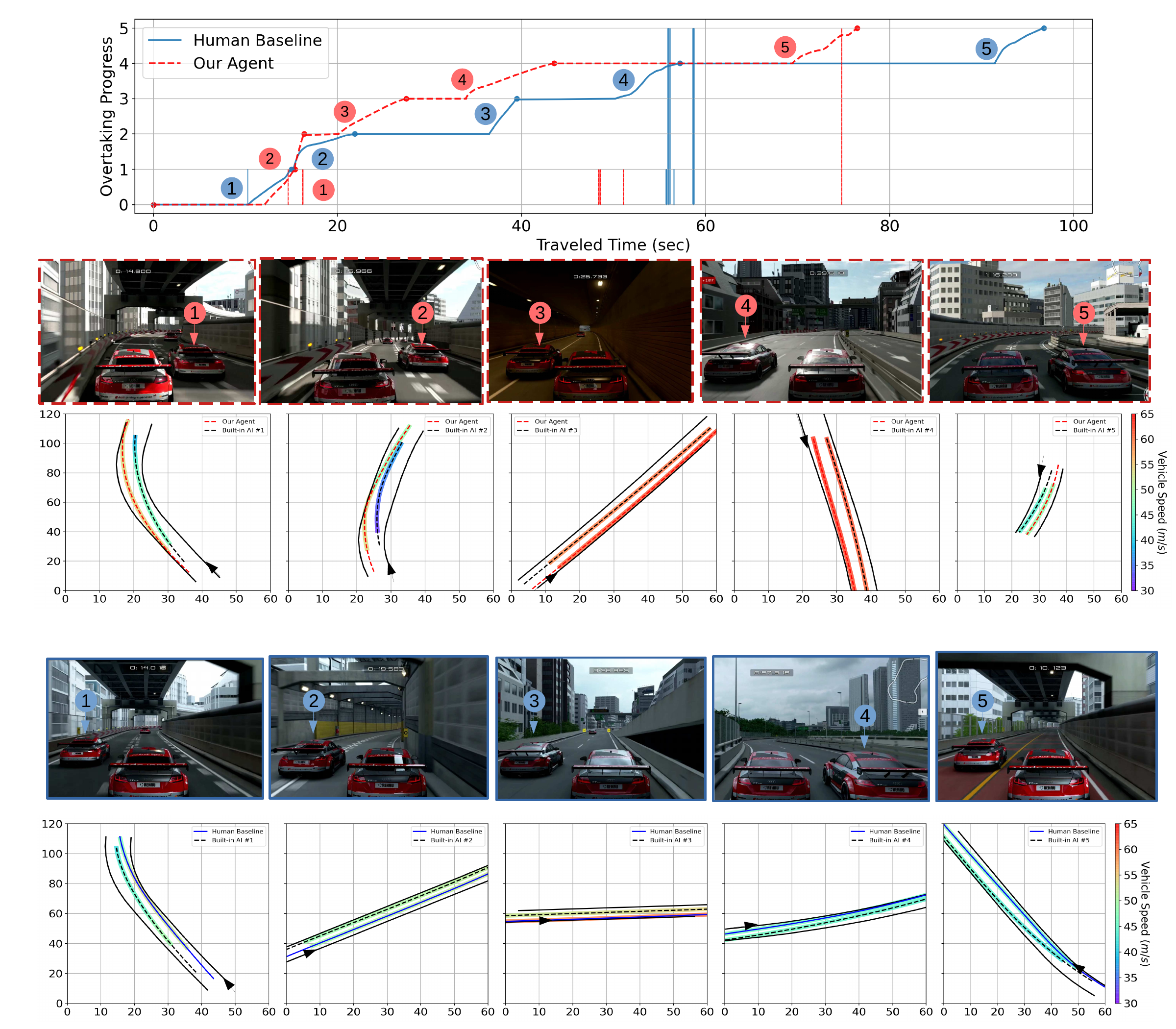}
    \caption{
    \textit{Top:} A comparison of the overtaking progress between our agent and an experienced human driver. 
    The long vertical straight lines indicate car collisions and the short vertical lines show wall collisions.
    \textit{Screenshots:} A visualization of five overtaking moments by our agent and the human player. 
    \textit{Trajectories:} The overtaking trajectories learned by our agent, the trajectories executed by the built-in game AI, and the overtaking trajectories performed by the human expert. 
    }
     \label{fig:overtaking_comparison}
\end{figure*}

To understand the overtaking strategy learned by our approach, 
we conduct a detailed analysis of the executed overtaking trajectory.
We compare the fastest trajectory executed by our agent (Agent3) with the fastest trajectory 
performed by the human player, both experiments are conducted using setting A.
Fig.~\ref{fig:overtaking_comparison} (\textit{Top}) shows a direct comparison of the overtaking progress between our agent and the human expert. 
In this comparison, it takes our agent less time to overtake all 5 front cars than that of the human driver.
However, our agent drives at high-speed which leads to more collisions with its opponents and wall.
Overall, our agent shows a comparable overtaking performance against the human expert. 

The trajectory plots in Fig.~\ref{fig:overtaking_comparison} (\textit{Middle}) show five overtaking trajectories (red dashed lines) performed by our agent and the trajectories (black solid lines) executed by the built-in game AI. 
The speeds are colored according to the color bar on the right. 
Our agent can maintain high-speed driving during the overtaking. 
Besides, our agent demonstrates different overtaking strategies in different driving scenarios, 
depending on both the track segment and the opponents' driving strategy. 
For example, the first and the second overtaking occurred consecutively on a difficult track segment,
which has a sharp turn. 
Our agent learns to drive along the outer side of the track at high speed 
since it has more free space.  
In addition, our agent manages to overtake its opponents on straight segments of the track. 
The screenshots provide a visualization of five different overtaking moments. 

As a comparison, the plots on the bottom show the overtaking trajectories 
performed by the human player. 
In summary, the human player can also overtake all the front vehicles,
but drives trajectories that are strategically different from those learned by our agent.  
For example, in the first overtaking segment, the human player took the inner side 
of the track, and hence, has to largely decrease its speed when entering the curve.  
Similarly, the human player can also perform overtaking on straight segments of the track
by simply speeding up the vehicle.

%% file: tables/training_hyperparameters.tex
\begin{table}[h]
\caption{Hyperparameters}
\label{tab-training-hyperparameters}
\begin{center}
\begin{small}
\setlength{\tabcolsep}{4pt}
\begin{tabular}{c|c}
\toprule
\rowcolor{Gray}
Hyperparameter & Value   \\
\midrule
{Neural network structure (MLP)}  & { 2 $\times$ [256, ReLU]} \\
\midrule
Mini-batch size & 4,096 \\
\midrule
Replay buffer size & $1 \times 10^6$ \\
\midrule
Start step (a trick to improve exploration) & $4 \times 10^4$ \\
\midrule
Learning rate & 0.001 \\
\midrule
Exponential discount factor & 0.99 \\
\midrule
Episode steps (under \SI{10}{\hertz} sampling rate) & 1,000 \\
\midrule
Total steps per epoch (20 cars in parallel) & 20,000 \\
\bottomrule
\end{tabular}
\end{small}
\end{center}
\end{table}

%% file: tables/baseline_comparison_result.tex
{\color{blue}

\begin{table}[h]
\caption{A baseline comparison for the single-car race. }
\label{tab-baseline-comparison}
\begin{center}
\begin{small}
\setlength{\tabcolsep}{4pt}
\begin{tabular}{m{4em}||c|c|c|c|c|c}
\toprule
\rowcolor{Gray}
  & {Built-in AI} & {BC} & {GAIL} & {SAC} & {TD3} & {PlaNet}\\
\midrule
 {Lap Time~(\si{\second})} & {86.9} & {108.0} & {144.9} & {80.1} & {80.8} & {89.0} \\
\midrule
 {Average Speed (\si{\kilo\metre\per\hour})} & {184.4} & {146.8} & {109.4} &  {198.2} & {196.3} & {178.2} \\
\bottomrule
\end{tabular}
\end{small}
\end{center}
\end{table}

}

%% file: sections/conclusion.tex
\section{CONCLUSION}
In this work, we proposed the usage of curriculum reinforcement learning to tackle high-speed autonomous race-car overtaking in Gran Turismo Sport. 
We demonstrated the advantages of curriculum RL over standard RL in autonomous overtaking, including better sample efficiency and overtaking performance. 
The learned overtaking policy outperforms the built-in model-based game AI and achieves comparable performance with an experienced human driver. 

Our empirical analysis suggests that complex tasks that are difficult to solve from scratch can be first sequenced into a curriculum and, then, be solved more efficiently with a stage-by-stage learning procedure. 
The proposed approach has limitations in terms of scalability and generalizability. In particular, the learned control policies are validated only in simulation and restricted to apply to a single track/car combination.
Nevertheless, the method presented in this paper can serve as a step towards developing more practical autonomous-driving systems in the real world. 

%% file: sections/acknowledgments.tex
\section{Acknowledgments}
We thank Kenta Kawamoto and Florian Fuchs from Sony AI Tokyo and Zurich, respectively, for their help and fruitful discussions.